\documentclass[runningheads]{llncs}
\usepackage[utf8]{inputenc}
\usepackage{graphicx}
\usepackage{amssymb,amsmath,mathtools}
\usepackage{enumitem}
\usepackage{hyperref}
\usepackage{pgfplots}
\usepackage{tikz}
\usepackage{csquotes}

\newcommand{\tsum}{\sum\nolimits}

\newcommand{\snorm}[1]{\left\lVert\smash{#1}\mathstrut\right\rVert}

\newcommand{\Var}{\ensuremath{\operatorname{Var}}}
\newcommand{\SSE}{\ensuremath{S}}
\newcommand{\weight}{\ensuremath{n}}

\newcommand{\CF}{\ensuremath{\mathrm{CF}}}
\newcommand{\CFBI}{\ensuremath{\CF^{\text{\tiny B\kern-.5pt I\kern-.5pt R\kern-.5pt C\kern-.5pt H\kern-.5pt}}}}
\newcommand{\CFBE}{\ensuremath{\CF^{\text{\tiny B\kern-.5pt E\kern-.5pt T\kern-.5pt U\kern-.5pt L\kern-.5pt A\kern-.5pt}}}}
\newcommand{\AB}{\ensuremath{\mathit{A\kern-1pt B}}}

\newcommand{\refsec}[1]{Section~\ref{#1}}
\newcommand{\reffig}[1]{Fig.~\ref{#1}}

\newcommand{\refeqn}[1]{Eq.~\eqref{#1}}

\newenvironment{texttabbing}
  {\setlength{\topsep}{0pt plus .1ex}%
   \setlength{\partopsep}{0pt plus .1ex}%
   \tabbing}
  {\endtabbing}

\makeatletter
\newcommand{\oset}[3][0ex]{%
  \mathrel{\mathop{#3}\limits^{
    \vbox to#1{\kern-2\ex@
    \hbox{$\scriptstyle#2$}\vss}}}}
\makeatother
\newcommand\Warning{%
\makebox[.8em][c]{%
\makebox[0pt][c]{\raisebox{.2ex}{\tiny!}}%
\makebox[0pt][c]{\color{red}\small$\boldsymbol\bigtriangleup$}}}%
\newcommand{\badminus}{\ensuremath{\smash{\oset[.15ex]{\scalebox{.8}{\Warning}}{-}}}}

\usepgfplotslibrary{colorbrewer}
\pgfplotsset{cycle list/Paired-12}
\pgfplotsset{
  compat=1.14,
  title style={font=\scriptsize, yshift=-.066em, inner sep=0pt},
  every tick label/.append style={font=\tiny, inner sep=0pt},
  every axis label/.append style={font=\tiny, inner sep=0pt},
  every axis legend/.append style={font=\tiny, inner sep=0pt},
  cycle list/.define={my marks}{
    every mark/.append style={solid,fill=\pgfkeysvalueof{/pgfplots/mark list fill}},mark=+\\
    every mark/.append style={solid,fill=\pgfkeysvalueof{/pgfplots/mark list fill}},mark=+\\
    every mark/.append style={solid,fill=\pgfkeysvalueof{/pgfplots/mark list fill}},mark=x,densely dotted,very thick\\
    every mark/.append style={solid,fill=\pgfkeysvalueof{/pgfplots/mark list fill}},mark=+\\  
    every mark/.append style={solid,fill=\pgfkeysvalueof{/pgfplots/mark list fill}},mark=+\\
    every mark/.append style={solid,fill=\pgfkeysvalueof{/pgfplots/mark list fill}},mark=x,densely dotted,very thick\\  
  },
  cycle list/.define={my colors}{Paired-D,Paired-B,Paired-A,Paired-H,Paired-F,Paired-E,Paired-J},
  cycle list/.define={my marks vertical}{
    every mark/.append style={solid,fill=\pgfkeysvalueof{/pgfplots/mark list fill}},mark=+\\
    every mark/.append style={solid,fill=\pgfkeysvalueof{/pgfplots/mark list fill}},mark=+\\
    every mark/.append style={solid,fill=\pgfkeysvalueof{/pgfplots/mark list fill}},mark=+\\
    every mark/.append style={solid,fill=\pgfkeysvalueof{/pgfplots/mark list fill}},mark=+\\  
    every mark/.append style={solid,fill=\pgfkeysvalueof{/pgfplots/mark list fill}},mark=x,densely dotted,very thick\\
    every mark/.append style={solid,fill=\pgfkeysvalueof{/pgfplots/mark list fill}},mark=x,densely dotted,very thick\\  
  },
  cycle list/.define={my colors vertical}{Paired-D,Paired-H,Paired-B,Paired-F,Paired-A,Paired-E,Paired-J},
  cycle multiindex* list={
    my marks\nextlist
    my colors\nextlist
  },
  cycle list/.define={buildtime marks}{
    every mark/.append style={solid,fill=\pgfkeysvalueof{/pgfplots/mark list fill}},mark=+\\
    every mark/.append style={solid,fill=\pgfkeysvalueof{/pgfplots/mark list fill}},mark=x,densely dotted\\
    every mark/.append style={solid,fill=\pgfkeysvalueof{/pgfplots/mark list fill}},mark=o,densely dashed\\
    every mark/.append style={solid,fill=\pgfkeysvalueof{/pgfplots/mark list fill}},mark=+\\  
    every mark/.append style={solid,fill=\pgfkeysvalueof{/pgfplots/mark list fill}},mark=x,densely dotted\\
    every mark/.append style={solid,fill=\pgfkeysvalueof{/pgfplots/mark list fill}},mark=o,densely dashed\\  
  },
  cycle list/.define={buildtime colors}{Paired-D,Paired-D,Paired-D,Paired-B,Paired-B,Paired-B,BuGn-K},
  legend cell align=left,
  every axis plot/.style={thick},
  legend style={row sep=-1pt, column sep=1pt, inner sep=1pt},
  x tick label style={yshift=-.5ex},
  y tick label style={xshift=-.5ex},
  y label style={yshift=.5ex, inner sep=0pt},
  x label style={yshift=-.7ex, inner sep=0pt},
  tick scale binop={\!\cdot\!},
}
\pgfkeys{/pgf/number format/.cd,sci generic={mantissa sep={\!\cdot\!},exponent={10^{#1}}}}

\newcommand{\plotGridLike}[2]{
	\begin{tikzpicture}
		\pgfplotsset{every axis/.append style={
			tick style={line width=0.8pt}}}
		\begin{axis}[
			legend pos=south west,
			legend columns = 3,
			width=#1,
			height=#2,
			title={},%
			xlabel={Scale of the Dataset},
			ylabel={Mean Log-Likelihood},
			xmin=0.08,
			xmax=1.02,
			ymin=-7.67528685895765,
			ymax=-7.5030215414044745,
		]
			\addplot coordinates {(0.1,-7.586497561316588) (0.2,-7.587065593617552) (0.4,-7.586017792447593) (0.6,-7.592871254058317) (0.8,-7.591293736357063) (1.0,-7.588850691527297)};
			\addplot coordinates {(0.1,-7.585114367514629) (0.2,-7.578448295088429) (0.4,-7.580327999494473) (0.6,-7.586963813823978) (0.8,-7.589901747149247) (1.0,-7.591807594001291)};
			\addplot coordinates {(0.1,-7.529652286595249) (0.2,-7.528139142557262) (0.4,-7.5253168462524345) (0.6,-7.53302988160779) (0.8,-7.534435501602867) (1.0,-7.530563511761593)};
			\addplot coordinates {(0.1,-7.603486821527943) (0.2,-7.608103887651907) (0.4,-7.603435790643934) (0.6,-7.604378457723309) (0.8,-7.602376221445459) (1.0,-7.607768102153419)};
			\addplot coordinates {(0.1,-7.520974287819227) (0.2,-7.520895546924048) (0.4,-7.517376984533906) (0.6,-7.5193496631906065) (0.8,-7.519048107518346) (1.0,-7.519399625105776)};
			\addplot coordinates {(0.1,-7.603486821527906) (0.2,-7.608103887652053) (0.4,-7.603435790643601) (0.6,-7.60437845772283) (0.8,-7.602376221445134) (1.0,-7.6077681021527885)};
			\legend{
				{BIRCH IGMM},
				{BETULA IGMM},
				{BETULA DGMM},
				{Stable IGMM},
				{Stable DGMM},
				{Textbook IGMM}
			}
		\end{axis}
	\end{tikzpicture}
}
\newcommand{\plotGridTime}[2]{
	\begin{tikzpicture}
		\pgfplotsset{every axis/.append style={
			tick style={line width=0.8pt}}}
		\begin{axis}[
			legend pos=north west,
			width=#1,
			height=#2,
			title={},%
			xlabel={Scale of the Dataset},
			ylabel={Mean Runtime [s]},
			xmin=0.08,
			xmax=1.02,
			ymin=0,
			ymax=679.8468777777777,
		]
			\addplot coordinates {(0.1,4.758333333333333) (0.2,5.995333333333332) (0.4,7.18988888888889) (0.6,8.933555555555555) (0.8,11.004222222222224) (1.0,13.471333333333336)};
			\addplot coordinates {(0.1,47.4430) (0.2,93.68555555555556) (0.4,204.61222222222228) (0.6,331.81222222222225) (0.8,456.30888888888893) (1.0,613.7343333333335)};
			\addplot coordinates {(0.1,51.3060) (0.2,101.87677777777777) (0.4,223.82788888888888) (0.6,351.93477777777775) (0.8,477.21733333333326) (1.0,618.3638888888889)};
			\addplot coordinates {(0.1,3.5340) (0.2,4.806777777777777) (0.4,5.872555555555556) (0.6,7.65988888888889) (0.8,9.818777777777776) (1.0,11.439777777777776)};
			\addplot coordinates {(0.1,4.745555555555556) (0.2,6.232444444444444) (0.4,7.16311111111111) (0.6,8.764666666666666) (0.8,10.835333333333334) (1.0,13.228333333333336)};
			\addplot coordinates {(0.1,49.60688888888888) (0.2,97.51433333333331) (0.4,214.27466666666663) (0.6,341.44644444444444) (0.8,463.14366666666674) (1.0,584.1407777777779)};
			\legend{
				{BIRCH IGMM},
				{BETULA IGMM},
				{BETULA DGMM},
				{Stable IGMM},
				{Stable DGMM},
				{Textbook IGMM}
			}
		\end{axis}
	\end{tikzpicture}
}

\newcommand{\plotRandomLike}[2]{
	\begin{tikzpicture}
		\pgfplotsset{every axis/.append style={
			tick style={line width=0.8pt}}}
		\begin{axis}[
			legend pos=south west,
			legend columns = 3,
			width=#1,
			height=#2,
			title={},%
			xlabel={Scale of the Dataset},
			ylabel={Mean Log-Likelihood},
			xmin=0.08,
			xmax=1.02,
			ymin=-7.410699556396695,
			ymax=-7.342163635756506,
		]
			\addplot coordinates {(0.1,-7.379716096365538) (0.2,-7.3839403495705875) (0.4,-7.382105500560621) (0.6,-7.387908097268293) (0.8,-7.383891281485207) (1.0,-7.383816342292019)};
			\addplot coordinates {(0.1,-7.380958545672952) (0.2,-7.3821665612666525) (0.4,-7.388928180806098) (0.6,-7.388422272270705) (0.8,-7.392806812664952) (1.0,-7.388816891282942)};
			\addplot coordinates {(0.1,-7.354360658483674) (0.2,-7.356235011533352) (0.4,-7.357006440395466) (0.6,-7.36200487330647) (0.8,-7.360549082703214) (1.0,-7.359852378853683)};
			\addplot coordinates {(0.1,-7.381212346028412) (0.2,-7.384094950688374) (0.4,-7.386464567436527) (0.6,-7.392296264791278) (0.8,-7.390524762505293) (1.0,-7.386691226232347)};
			\addplot coordinates {(0.1,-7.349620187999448) (0.2,-7.352194071362116) (0.4,-7.356125631216442) (0.6,-7.359908805087604) (0.8,-7.357602818323183) (1.0,-7.356385391395287)};
			\addplot coordinates {(0.1,-7.381212346028602) (0.2,-7.384094950688389) (0.4,-7.386464567436457) (0.6,-7.392296264791254) (0.8,-7.39052476250526) (1.0,-7.386691226232325)};
			\legend{
				{BIRCH IGMM},
				{BETULA IGMM},
				{BETULA DGMM},
				{Stable IGMM},
				{Stable DGMM},
				{Textbook IGMM}
			}
		\end{axis}
	\end{tikzpicture}
}
\newcommand{\plotRandomTime}[2]{
	\begin{tikzpicture}
		\pgfplotsset{every axis/.append style={
			tick style={line width=0.8pt}}}
		\begin{axis}[
			legend pos = north west,
			width=#1,
			height=#2,
			title={},%
			xlabel={Scale of the Dataset},
			ylabel={Mean Runtime [s]},
			xmin=0.08,
			xmax=1.02,
			ymin=0,
			ymax=655.8852,
		]
			\addplot coordinates {(0.1,4.1550) (0.2,5.5040) (0.4,5.9730) (0.6,7.9670) (0.8,10.0990) (1.0,12.0530)};
			\addplot coordinates {(0.1,44.0940) (0.2,86.7930) (0.4,191.5440) (0.6,306.1680) (0.8,423.5290) (1.0,585.5130)};
			\addplot coordinates {(0.1,47.4530) (0.2,93.4330) (0.4,211.2010) (0.6,333.8010) (0.8,448.2840) (1.0,596.4450)};
			\addplot coordinates {(0.1,3.1060) (0.2,4.1580) (0.4,5.1840) (0.6,7.2800) (0.8,8.9870) (1.0,11.3670)};
			\addplot coordinates {(0.1,4.1030) (0.2,5.7320) (0.4,6.0020) (0.6,8.6040) (0.8,10.1530) (1.0,11.8360)};
			\addplot coordinates {(0.1,45.8320) (0.2,87.9940) (0.4,198.6620) (0.6,322.0540) (0.8,434.3690) (1.0,575.8620)};
			\legend{
				{BIRCH IGMM},
				{BETULA IGMM},
				{BETULA DGMM},
				{Stable IGMM},
				{Stable DGMM},
				{Textbook IGMM}
			}
		\end{axis}
	\end{tikzpicture}
}
\newcommand{\plotShiftLike}[2]{
    \vspace{-1em}
	\begin{tikzpicture}
		\pgfplotsset{every axis/.append style={
			tick style={line width=0.8pt}}}
		\begin{semilogxaxis}[
            legend columns = 2,
            cycle multiindex* list={
    			my marks vertical\nextlist
    			my colors vertical\nextlist
  			},
			legend pos=south west,
			width=#1,
			height=#2,
			title={},%
			xlabel={Relative distance of Clusters},
			ylabel={Mean Log-Likelihood},
			xmin=0.08,
			xmax=111111111.11111103,
			ymin=-85,
			ymax=0,
		]
			\addplot coordinates {(0.1,-4.332742914276524) (0.2,-4.389896408120237) (0.3,-4.475599762843116) (0.4,-4.577826685229074) (0.5,-4.683480412567318) (0.6,-4.781980865743498) (0.7,-4.867130634815537) (0.8,-4.936502395682508) (0.9,-4.9902142899746496) (1.0,-5.029941794903774) (2.0,-5.108908492534799) (3.0,-5.109090068996203) (4.0,-5.109090068997952) (5.0,-5.1090900689979515) (7.0,-5.1090900689979515) (10.0,-5.1090900689979515) (20.0,-5.109090068997948) (50.0,-5.109090068997946) (70.0,-5.109090068997945) (100.0,-5.109090068997953) (200.0,-5.109090068997946) (500.0,-5.109090068997941) (1000.0,-5.109090068997947) (2000.0,-5.109090068997976) (5000.0,-5.109090068997959) (10000.0,-5.109090068997955) (20000.0,-5.109090068997956) (50000.0,-5.109090068997963) (100000.0,-5.109090068998343) (200000.0,-5.10909006902421) (500000.0,-5.109090072014743) (1000000.0,-5.109090128544099) (2000000.0,-5.109091526625108) (5000000.0,-5.109428155356612) (10000000.0,-5.175720848951358) (20000000.0,-18.85620563302792) (50000000.0,-78.1332573649195) (100000000.0,-266.71280849572577)};
			\addplot coordinates {(0.1,-4.332590137363762) (0.2,-4.389743117518673) (0.3,-4.475449593341245) (0.4,-4.5776905662464715) (0.5,-4.683364263476929) (0.6,-4.781890816878613) (0.7,-4.867070592541834) (0.8,-4.936466081707031) (0.9,-4.990194318435865) (1.0,-5.029931710678691) (2.0,-5.108908491617831) (3.0,-5.109090068996204) (4.0,-5.109090068997946) (5.0,-5.10909006899795) (7.0,-5.109090068997955) (10.0,-5.109090068997942) (20.0,-5.109090068997963) (50.0,-5.10909006899796) (70.0,-5.109090068997992) (100.0,-5.109090068997982) (200.0,-5.109090068997988) (500.0,-5.109090068997963) (1000.0,-5.109090068997995) (2000.0,-5.109090068997965) (5000.0,-5.109090068998171) (10000.0,-5.109090069002543) (20000.0,-5.109090069092968) (50000.0,-5.1090900728573) (100000.0,-5.109090102622675) (200000.0,-5.1090903760912125) (500000.0,-5.109132921243606) (1000000.0,-5.109681772078066) (2000000.0,-5.122796994337133) (5000000.0,-9.810394719834123) (10000000.0,-25.19945799544076) (20000000.0,-50.24786232344607) (50000000.0,-53.038864155363676) (100000000.0,-26.46746919959639)};
			\addplot coordinates {(0.1,-4.3327456001711715) (0.2,-4.389896574566641) (0.3,-4.47559842481691) (0.4,-4.5778271104141925) (0.5,-4.683477063823166) (0.6,-4.781980681459746) (0.7,-4.867133487034795) (0.8,-4.93650332842326) (0.9,-4.9902158538153385) (1.0,-5.029942297486898) (2.0,-5.108908493834519) (3.0,-5.109090068996207) (4.0,-5.109090068997952) (5.0,-5.1090900689979515) (7.0,-5.109090068997952) (10.0,-5.1090900689979515) (20.0,-5.10909006899795) (50.0,-5.1090900689979515) (70.0,-5.109090068997945) (100.0,-5.10909006899795) (200.0,-5.1090900689979435) (500.0,-5.109090068997959) (1000.0,-5.1090900689979435) (2000.0,-5.109090068997942) (5000.0,-5.109090068997942) (10000.0,-5.109090068997918) (20000.0,-5.109090068998003) (50000.0,-5.109090068997936) (100000.0,-5.109090068997952) (200000.0,-5.10909006899796) (500000.0,-5.109090068997888) (1000000.0,-5.109090068998203) (2000000.0,-5.1090900689981975) (5000000.0,-5.1090900689971805) (10000000.0,-5.109090068997338) (20000000.0,-5.109090068996243) (50000000.0,-5.109090069009845) (100000000.0,-5.109090069003747)};
			\addplot coordinates {(0.1,-4.277411526873919) (0.2,-4.3382947657258795) (0.3,-4.42447365016565) (0.4,-4.521760181849709) (0.5,-4.617821483466874) (0.6,-4.7041305108606615) (0.7,-4.7760136795173285) (0.8,-4.832092951384199) (0.9,-4.873335118174752) (1.0,-4.901990432401087) (2.0,-4.948279382646284) (3.0,-4.9483100661254475) (4.0,-4.9483100661254475) (5.0,-4.9483100661254475) (7.0,-4.948310066125447) (10.0,-4.948310066125446) (20.0,-4.948310066125448) (50.0,-4.948310066125444) (70.0,-4.948310066125441) (100.0,-4.948310066125446) (200.0,-4.948310066125446) (500.0,-4.948310066125441) (1000.0,-4.948310066125456) (2000.0,-4.948310066125453) (5000.0,-4.948310066125488) (10000.0,-4.94831006612546) (20000.0,-4.94831006612551) (50000.0,-4.948310066125498) (100000.0,-4.948310066125517) (200000.0,-4.948310066125473) (500000.0,-4.94831006612549) (1000000.0,-4.948310066125823) (2000000.0,-4.9483100661253765) (5000000.0,-4.948310066124288) (10000000.0,-4.948310066125766) (20000000.0,-4.948310066124941) (50000000.0,-4.948310066132784) (100000000.0,-4.948310066304338)};
			\addplot coordinates {(0.1,-4.332590137363764) (0.2,-4.389743117518673) (0.3,-4.475449593341247) (0.4,-4.57769056624647) (0.5,-4.683364263476929) (0.6,-4.781890816878613) (0.7,-4.867070592541834) (0.8,-4.936466081707028) (0.9,-4.990194318435865) (1.0,-5.029931710678691) (2.0,-5.108908491617832) (3.0,-5.109090068996203) (4.0,-5.1090900689979515) (5.0,-5.10909006899795) (7.0,-5.10909006899795) (10.0,-5.10909006899795) (20.0,-5.109090068997948) (50.0,-5.10909006899795) (70.0,-5.109090068997954) (100.0,-5.109090068997955) (200.0,-5.109090068997952) (500.0,-5.109090068997966) (1000.0,-5.109090068997978) (2000.0,-5.10909006899795) (5000.0,-5.109090068997993) (10000.0,-5.109090068997939) (20000.0,-5.109090068997992) (50000.0,-5.109090068997981) (100000.0,-5.109090068997928) (200000.0,-5.109090068997974) (500000.0,-5.109090068997925) (1000000.0,-5.109090068998187) (2000000.0,-5.109090068998216) (5000000.0,-5.109090068997156) (10000000.0,-5.109090068997366) (20000000.0,-5.109090068996403) (50000000.0,-5.109090069010176) (100000000.0,-5.109090069004979)};
			\addplot coordinates {(0.1,-4.27738179069173) (0.2,-4.337836258443398) (0.3,-4.423874939911664) (0.4,-4.521036126675435) (0.5,-4.617207997436771) (0.6,-4.703691011912472) (0.7,-4.775737261478383) (0.8,-4.831945826837863) (0.9,-4.8732541288417695) (1.0,-4.901952903293818) (2.0,-4.948279378853225) (3.0,-4.948310066125444) (4.0,-4.948310066125447) (5.0,-4.948310066125447) (7.0,-4.948310066125446) (10.0,-4.948310066125446) (20.0,-4.948310066125448) (50.0,-4.948310066125438) (70.0,-4.948310066125444) (100.0,-4.948310066125441) (200.0,-4.948310066125447) (500.0,-4.94831006612546) (1000.0,-4.948310066125456) (2000.0,-4.948310066125488) (5000.0,-4.9483100661255115) (10000.0,-4.948310066125496) (20000.0,-4.948310066125531) (50000.0,-4.948310066125504) (100000.0,-4.948310066125493) (200000.0,-4.948310066125488) (500000.0,-4.948310066125479) (1000000.0,-4.948310066125871) (2000000.0,-4.948310066125368) (5000000.0,-4.948310066124252) (10000000.0,-4.9483100661257655) (20000000.0,-4.9483100661250665) (50000000.0,-4.9483100661332005) (100000000.0,-4.94831006613382)};
			\legend{
				{BIRCH IGMM},
				{Textbook IGMM},
				{BETULA IGMM},
				{BETULA DGMM},
				{Stable IGMM},
				{Stable DGMM},
			}
		\end{semilogxaxis}
		\pgfresetboundingbox
		\path
					(current axis.south west)+(0,-0.2)
		rectangle 	(current axis.north east);
	\end{tikzpicture}
}

\newcommand{\plotRandomBuild}[2]{
	\begin{tikzpicture}
		\pgfplotsset{every axis/.append style={
			tick style={line width=0.8pt}}}
		\begin{axis}[
			legend pos=north west,
			legend columns = 3,
			cycle multiindex* list={
				buildtime colors
					\nextlist
				buildtime marks
					\nextlist
			},
			width=#1,
			height=#2,
			title={},%
			xlabel={Size of the data set},
			ylabel={Avg. CF-Tree build time [s]},
			xmin=80000,
			xmax=1020000,
			ymin=-0,
			ymax=14.97309,
			scaled x ticks = false,
		]
			\addplot coordinates {(100000,1.5303) (200000,2.4280) (400000,2.8857) (600000,3.0032) (800000,3.4885) (1000000,5.3071)};
			\addplot coordinates {(100000,2.8255) (200000,3.0876) (400000,6.3304) (600000,4.9958) (800000,5.5081) (1000000,9.4133)};
			\addplot coordinates {(100000,2.9868) (200000,2.3629) (400000,5.0614) (600000,6.9183) (800000,5.2900) (1000000,8.9991)};
			\addplot coordinates {(100000,2.4660) (200000,3.6838) (400000,5.2675) (600000,5.0000) (800000,4.5038) (1000000,7.5953)};
			\addplot coordinates {(100000,5.5444) (200000,6.5188) (400000,9.8836) (600000,8.9019) (800000,9.1972) (1000000,13.4744)};
			\addplot coordinates {(100000,5.4204) (200000,7.4330) (400000,10.1969) (600000,10.4807) (800000,9.9199) (1000000,12.8669)};
			\legend{
				{BIRCH 2k leaves},
				{BIRCH 5k leaves},
				{BIRCH 8k leaves},
				{BETULA 2k leaves},
				{BETULA 5k leaves},
				{BETULA 8k leaves}
			}
		\end{axis}
	\end{tikzpicture}\vspace{-1em}
}
\newcommand{\plotCombTime}[2]{
	\begin{tikzpicture}
		\pgfplotsset{every axis/.append style={
			tick style={line width=0.8pt}}}
		\begin{loglogaxis}[
            legend pos=north west,
			legend style={draw=none},
            legend columns = 3,
            legend entries = {
				{BIRCH IGMM},
				{BETULA IGMM},
				{BETULA DGMM},
				{Textbook IGMM},
				{Stable IGMM},
				{Stable DGMM},
            },
            legend to name = BLeg,
			width=#1/2,
			height=#2,
			title={Grid},
			xlabel={Size of the data set [log]},
			ylabel={Mean Runtime [s, log]},
			xmin=90000,
			xmax=1100000,
			ymin=0,
			ymax=1050,
			scaled x ticks = false,
		]
			\addplot coordinates {(100000,4.758333333333333) (200000,5.995333333333332) (400000,7.18988888888889) (600000,8.933555555555555) (800000,11.004222222222224) (1000000,13.471333333333336)};
			\addplot coordinates {(100000,3.5340) (200000,4.806777777777777) (400000,5.872555555555556) (600000,7.65988888888889) (800000,9.818777777777776) (1000000,11.439777777777776)};
			\addplot coordinates {(100000,4.745555555555556) (200000,6.232444444444444) (400000,7.16311111111111) (600000,8.764666666666666) (800000,10.835333333333334) (1000000,13.228333333333336)};
			\addplot coordinates {(100000,49.60688888888888) (200000,97.51433333333331) (400000,214.27466666666663) (600000,341.44644444444444) (800000,463.14366666666674) (1000000,584.1407777777779)};
			\addplot coordinates {(100000,47.4430) (200000,93.68555555555556) (400000,204.61222222222228) (600000,331.81222222222225) (800000,456.30888888888893) (1000000,613.7343333333335)};
			\addplot coordinates {(100000,51.3060) (200000,101.87677777777777) (400000,223.82788888888888) (600000,351.93477777777775) (800000,477.21733333333326) (1000000,618.3638888888889)};			
		\end{loglogaxis}
    \end{tikzpicture}%
    \begin{tikzpicture}
		\pgfplotsset{every axis/.append style={
			tick style={line width=0.8pt}}}
		\begin{loglogaxis}[
			width=#1/2,
			height=#2,
			title={Random},
			xlabel={Size of the data set [log]},
			ylabel={},
			xmin=90000,
			xmax=1100000,
			ymin=0,
			ymax=1050,
			scaled x ticks = false,
		]
			\addplot coordinates {(100000,4.1550) (200000,5.5040) (400000,5.9730) (600000,7.9670) (800000,10.0990) (1000000,12.0530)};
			\addplot coordinates {(100000,3.1060) (200000,4.1580) (400000,5.1840) (600000,7.2800) (800000,8.9870) (1000000,11.3670)};
			\addplot coordinates {(100000,4.1030) (200000,5.7320) (400000,6.0020) (600000,8.6040) (800000,10.1530) (1000000,11.8360)};
			\addplot coordinates {(100000,45.8320) (200000,87.9940) (400000,198.6620) (600000,322.0540) (800000,434.3690) (1000000,575.8620)};
			\addplot coordinates {(100000,44.0940) (200000,86.7930) (400000,191.5440) (600000,306.1680) (800000,423.5290) (1000000,585.5130)};
			\addplot coordinates {(100000,47.4530) (200000,93.4330) (400000,211.2010) (600000,333.8010) (800000,448.2840) (1000000,596.4450)};
		\end{loglogaxis}
	\end{tikzpicture}
    \ref*{BLeg}\vspace{-1em}
}

\newcommand{\plotCombLike}[2]{
	\begin{tikzpicture}
		\pgfplotsset{every axis/.append style={
			tick style={line width=0.8pt}}}
		\begin{axis}[
			legend pos=south west,
			legend style={draw=none},
            legend columns = 3,
            legend entries = {
				{BIRCH IGMM},
				{BETULA IGMM},
				{BETULA DGMM},
				{Textbook IGMM},
				{Stable IGMM},
				{Stable DGMM},
            },
            legend to name = BLeg2,
			width=#1/2,
			height=#2/1.5,
			title={Grid},%
			xlabel={Size of the data set},
			ylabel={Mean Log-Likelihood},
			xmin=80000,
			xmax=1020000,
			ymin=-7.65,
			ymax=-7.28,
			scaled x ticks = false,
		]
			\addplot coordinates {(100000,-7.586497561316588) (200000,-7.587065593617552) (400000,-7.586017792447593) (600000,-7.592871254058317) (800000,-7.591293736357063) (1000000,-7.588850691527297)};
			\addplot coordinates {(100000,-7.585114367514629) (200000,-7.578448295088429) (400000,-7.580327999494473) (600000,-7.586963813823978) (800000,-7.589901747149247) (1000000,-7.591807594001291)};
			\addplot coordinates {(100000,-7.529652286595249) (200000,-7.528139142557262) (400000,-7.5253168462524345) (600000,-7.53302988160779) (800000,-7.534435501602867) (1000000,-7.530563511761593)};
			\addplot coordinates {(100000,-7.603486821527906) (200000,-7.608103887652053) (400000,-7.603435790643601) (600000,-7.60437845772283) (800000,-7.602376221445134) (1000000,-7.6077681021527885)};
			\addplot coordinates {(100000,-7.603486821527943) (200000,-7.608103887651907) (400000,-7.603435790643934) (600000,-7.604378457723309) (800000,-7.602376221445459) (1000000,-7.607768102153419)};
			\addplot coordinates {(100000,-7.520974287819227) (200000,-7.520895546924048) (400000,-7.517376984533906) (600000,-7.5193496631906065) (800000,-7.519048107518346) (1000000,-7.519399625105776)};
		\end{axis}
    \end{tikzpicture}
    \begin{tikzpicture}
		\pgfplotsset{every axis/.append style={
			tick style={line width=0.8pt}}}
		\begin{axis}[
			legend pos=south west,
			width=#1/2,
			height=#2/1.5,
			title={Random},%
			xlabel={Size of the data set},
			ylabel={},
			xmin=80000,
			xmax=1020000,
			ymin=-7.65,
			ymax=-7.28,
			scaled x ticks = false,
		]
			\addplot coordinates {(100000,-7.379716096365538) (200000,-7.3839403495705875) (400000,-7.382105500560621) (600000,-7.387908097268293) (800000,-7.383891281485207) (1000000,-7.383816342292019)};
			\addplot coordinates {(100000,-7.380958545672952) (200000,-7.3821665612666525) (400000,-7.388928180806098) (600000,-7.388422272270705) (800000,-7.392806812664952) (1000000,-7.388816891282942)};
			\addplot coordinates {(100000,-7.354360658483674) (200000,-7.356235011533352) (400000,-7.357006440395466) (600000,-7.36200487330647) (800000,-7.360549082703214) (1000000,-7.359852378853683)};
			\addplot coordinates {(100000,-7.381212346028602) (200000,-7.384094950688389) (400000,-7.386464567436457) (600000,-7.392296264791254) (800000,-7.39052476250526) (1000000,-7.386691226232325)};
			\addplot coordinates {(100000,-7.381212346028412) (200000,-7.384094950688374) (400000,-7.386464567436527) (600000,-7.392296264791278) (800000,-7.390524762505293) (1000000,-7.386691226232347)};
			\addplot coordinates {(100000,-7.349620187999448) (200000,-7.352194071362116) (400000,-7.356125631216442) (600000,-7.359908805087604) (800000,-7.357602818323183) (1000000,-7.356385391395287)};
		\end{axis}
    \end{tikzpicture}
    \ref*{BLeg2}\vspace{-1em}
}

\newcommand{\plotTrafficCombined}[2]{
	\begin{tikzpicture}
		\pgfplotsset{every axis/.append style={
			tick style={line width=0.8pt}}}
		\begin{semilogxaxis}[
			legend pos=south west,
			legend style={draw=none},
            legend columns = 3,
            legend entries = {
				{BIRCH IGMM},
				{BETULA IGMM},
				{BETULA DGMM},
				{\enskip},
				{Stable IGMM},
				{Stable DGMM},
            },
            legend to name = BLeg3,
			width=#1/2,
			height=#2/1,
			title={Log-Likelihood},%
			xlabel={Number of Clusters},
			ylabel={Mean Log-Likelihood},
			xmin=9,
			xmax=1100,
			ymin=-26.1,
            ymax=-24.8,
		]
			\addplot coordinates {(10.0,-25.722057294121278) (20.0,-25.58176836765743) (30.0,-25.469838390972228) (40.0,-25.445965008713284) (50.0,-25.427179825322213) (60.0,-25.403655477496876) (80.0,-25.386488260086608) (100.0,-25.380689920884162) (200.0,-25.37396936254549) (300.0,-25.37151134000045) (400.0,-25.37039361560902) (600.0,-25.371220620286397) (800.0,-25.384609139562567) (1000.0,-25.3838715676757)};
			\addplot coordinates {(10.0,-25.71151919874044) (20.0,-25.54911011113104) (30.0,-25.414192320496035) (40.0,-25.361625301184226) (50.0,-25.320090453537112) (60.0,-25.28679933961373) (80.0,-25.23721076508297) (100.0,-25.202301338189397) (200.0,-25.100766880313405) (300.0,-25.044704108659754) (400.0,-25.02174506781609) (600.0,-24.97606893472747) (800.0,-24.946042797892915) (1000.0,-24.92849970373563)};
			\addplot coordinates {(10.0,-25.64530602660735) (20.0,-25.44973096101223) (30.0,-25.363142313034864) (40.0,-25.31610172500373) (50.0,-25.282530383085817) (60.0,-25.251778811181197) (80.0,-25.20816440394794) (100.0,-25.171880802047376) (200.0,-25.068545528983456) (300.0,-25.01134261233212) (400.0,-24.97778346651608) (600.0,-24.94133576670433) (800.0,-24.919181613181003) (1000.0,-24.90446615695125)};
			\addplot[opacity=0] coordinates {(10,-25)}; %
			\addplot coordinates {(10.0,-25.611564769900998) (20.0,-25.460611428610893) (30.0,-25.374699794365558) (40.0,-25.32077336652824) (50.0,-25.258705844009285)};
			\addplot coordinates {(10.0,-25.596628854868086) (20.0,-25.41518859818881) (30.0,-25.331915090424882) (40.0,-25.282977811840915) (50.0,-25.23213528030951)};
		\end{semilogxaxis}
    \end{tikzpicture}
    \begin{tikzpicture}
		\pgfplotsset{every axis/.append style={
			tick style={line width=0.8pt}}}
		\begin{semilogxaxis}[
			legend pos=north east,
			width=#1/2,
			height=#2/1,
			title={Runtime},%
			xlabel={Number of Clusters},
			ylabel={Mean Runtime in [s]},
			xmin=9,
			xmax=1100,
			ymin=-100,
			ymax=1590,
		]
			\addplot coordinates {(10.0,17.1318) (20.0,21.2669) (30.0,27.4275) (40.0,28.6120) (50.0,33.3588) (60.0,39.4188) (80.0,45.8589) (100.0,62.0748) (200.0,108.6640) (300.0,144.9119) (400.0,203.1606) (600.0,276.7770) (800.0,377.8378) (1000.0,458.7977)};
			\addplot coordinates {(10.0,22.8575) (20.0,26.5428) (30.0,31.6078) (40.0,35.0840) (50.0,40.1089) (60.0,39.1095) (80.0,52.1486) (100.0,62.5927) (200.0,109.0590) (300.0,150.9514) (400.0,179.6384) (600.0,253.1825) (800.0,321.6491) (1000.0,389.4010)};
			\addplot coordinates {(10.0,23.4610) (20.0,27.3049) (30.0,32.7642) (40.0,35.6844) (50.0,39.9763) (60.0,44.9124) (80.0,50.6470) (100.0,62.4780) (200.0,108.9283) (300.0,149.1381) (400.0,199.3412) (600.0,261.3001) (800.0,368.2422) (1000.0,396.8545)};
			\pgfplotsset{cycle list shift=1}
			\addplot coordinates {(10.0,332.5992) (20.0,708.7274) (30.0,998.6746) (40.0,1271.1469) (50.0,1536.2219)};
			\addplot coordinates {(10.0,366.3812) (20.0,686.9695) (30.0,961.3283) (40.0,1232.0231) (50.0,1518.5891)};
        \end{semilogxaxis}
    \end{tikzpicture}
    \ref*{BLeg3}\vspace{-1em}
}

\hypersetup{pdfborderstyle={/S/U/W 0}}
\hypersetup{pdfborder={0 0 0}}%

\begin{document}
\title{BETULA: Numerically Stable CF-Trees\newline for BIRCH Clustering%
\thanks{Part of the work on this paper has been supported by Deutsche Forschungsgemeinschaft (DFG)
within the Collaborative Research Center SFB 876
``Providing Information by Resource-Constrained Analysis'', project A2 
}
}

\titlerunning{BETULA: Numerically Stable CF-Trees for BIRCH Clustering}

\author{Andreas Lang\orcidID{0000-0003-3212-5548} \and
Erich Schubert\orcidID{0000-0001-9143-4880}}
\authorrunning{A. Lang and E. Schubert}
\institute{TU Dortmund University, Dortmund, Germany 
\email{\{andreas.lang,erich.schubert\}@tu-dortmund.de}
}

\maketitle

\begin{abstract}
BIRCH clustering is a widely known approach for clustering,
that has influenced much subsequent research and commercial products.
The key contribution of BIRCH is the Clustering Feature tree (CF-Tree),
which is a compressed representation of the input data.
As new data arrives, the tree is eventually rebuilt to
increase the compression. Afterward, the leaves of the tree
are used for clustering. Because of the data compression, this
method is very scalable. 
The idea has been adopted for example for $k$-means, data stream, and density-based clustering. 

Clustering features used by BIRCH are simple summary statistics
that can easily be updated with new data: 
the number of points, the linear sums, and the sum of squared values.
Unfortunately, how the sum of squares is then used in BIRCH is prone to
catastrophic cancellation.

We introduce a replacement cluster feature that does not have this numeric problem, that
is not much more expensive to maintain, and which makes many
computations simpler and hence more
efficient. These cluster features can also easily be used in
other work derived from BIRCH, such as algorithms for streaming data.
In the experiments, we demonstrate the numerical problem
and compare the performance of the original algorithm compared
to the improved cluster features.

\end{abstract}

\section{Introduction}
The BIRCH algorithm \cite{DBLP:conf/sigmod/ZhangRL96,DBLP:journals/datamine/ZhangRL97,tr/wisc/Zhang97}
is a widely known cluster analysis approach, %
that won the 2006 SIGMOD Test of Time Award.
It scales well to big data even with limited resources
because it processes the data as a stream and aggregates it into a compact summary of the data.
BIRCH has inspired many subsequent works, such as
two-step clustering~\cite{DBLP:conf/kdd/ChiuFCWJ01}, %
data bubbles~\cite{DBLP:conf/sigmod/BreunigKKS01},
and stream clustering methods such as CluStream~\cite{DBLP:conf/vldb/AggarwalHWY03}
and DenStream \cite{DBLP:conf/sdm/CaoEQZ06}.
Clustering is the unsupervised learning task aimed at discovering potential
structure in a data set when no labeled data or pattern examples are available.
It is inherently underspecified and
subjective \cite{DBLP:journals/sigkdd/Estivill-Castro02,DBLP:journals/ibmrd/Bonner64}
and, unfortunately, also very difficult to evaluate. Instead, it is
best approached as explorative data analysis, generating
hypotheses about potential structures in the data, that afterward need
to be verified by some other procedure, which is domain-specific and
may require a domain expert to inspect the results. %
Many clustering algorithms and evaluation measures have been proposed
with unclear advantages of one over another.
Because many of the underlying problems (e.g., $k$-means clustering) are NP-hard,
we often use approximation techniques and great concern is directed at the scalability. %

Scalability is where the BIRCH algorithm shines. It is a multi-step procedure for
numerical data that first aggregates the data %
into a tree-based data
structure much smaller than the original data. This condensed representation is
then fed into a clustering method, which now is faster
because of the reduced size. The main contribution of BIRCH is a flexible logic
for aggregating the data so that an informative representation is retained
even when the size is reduced substantially.

When studying BIRCH closely, we noticed that it is susceptible to a numerical
problem known as ``catastrophic cancellation''.
This arises when two large and similar floating-point values are
subtracted: many bits of the significand cancel out,
and only few bits of valid result remain.
In this paper, we show how to avoid this numerical problem and
demonstrate that it can arise in real data even at low dimensionality.
We propose a replacement cluster feature tree (BETULA) that does not suffer from
this numeric problem while retaining all functionality.
Furthermore, it is often even easier to use. 
This structure can easily be integrated into most (if not all) derived methods,
in particular also for data streams.

\section{Related Work}

The BIRCH algorithm was presented at the SIGMOD conference \cite{DBLP:conf/sigmod/ZhangRL96},
then expanded in a journal version \cite{DBLP:journals/datamine/ZhangRL97}.
Still, both versions omit integral details of the algorithm
(e.g., Eqs.~\refeq{eq:d2-birch} to \refeq{eq:d4-birch} below to compute
distances using cluster features), which are found only in
their technical report \cite{tr/wisc/Zhang97} or their source code.
Nevertheless, the intriguing ideas of the clustering features and the CF-Tree inspired a plethora of subsequent work.
Bradley et al.{}~\cite{DBLP:conf/kdd/BradleyFR98} use the same ``clustering features'' as BIRCH,
but call them ``sub-cluster sufficient statistics''.
The CF-Tree has also been used for kernel density estimation~\cite{DBLP:conf/kdd/ZhangRL99},
with a threshold set on the variance to guarantee approximation quality.
In two-step clustering~\cite{DBLP:conf/kdd/ChiuFCWJ01}, BIRCH is extended to mixed data,
by adding histograms over the categorical variables.

Because BIRCH is sequentially inserting data points into
the CF-tree, the tree construction can be suspended at any time.
The leaves can then be processed with a clustering algorithm;
when new data arrives the tree construction is continued
and we trivially obtain a stream clustering algorithm~\cite{DBLP:journals/tkde/GantiGR01}.
CluStream~\cite{DBLP:conf/vldb/AggarwalHWY03} extends this idea with pyramidal time frames
to enable the clustering of parts of the data stream by integrating temporal information.
HPStream~\cite{DBLP:conf/vldb/AggarwalHWY04} extends CluStream to projected/subspace clustering.
DenStream~\cite{DBLP:conf/sdm/CaoEQZ06} uses clustering features for density-based stream clustering
to detect clusters of arbitrary shape (in contrast to earlier methods that focus on $k$-means-style clustering).
Breunig et al.~\cite{DBLP:conf/pkdd/BreunigKS00} adopt clustering features to perform hierarchical density-based 
OPTICS clustering~\cite{DBLP:conf/sigmod/AnkerstBKS99} on large data.
The ClusTree~\cite{DBLP:journals/kais/KranenABS11} combines %
R-trees with BIRCH clustering features to process data streams.
BICO~\cite{DBLP:conf/esa/FichtenbergerGSSS13} aims at improving the theoretical
foundations (and hence, performance guarantees) of BIRCH by combining it with
the concept of coresets. For this, it is necessary to add
reference points to the clustering features and use a strict radius threshold.

\section{BIRCH and BETULA}
In this section, we will describe the basic BIRCH tree building algorithm,
and introduce the changes made for BETULA to become numerically more reliable.

\subsection{BIRCH Clustering Features}

The central concept of BIRCH is a summary data structure known as
Clustering Features $\CFBI{=}(\vec{LS}, SS, N)$.
Each clustering feature represents $N$ data points, summarized
using the linear sum vector $\vec{LS}{\in}\mathbb{R}^d$ (with $LS_i{=}\sum_x x_i$),
the sum of squares $SS{\in}\mathbb{R}$ (originally not a vector,
but a scalar $SS{=}\sum_i \sum_x x_i^2$) and the count $N{\in}\mathbb{N}$.
The center of a clustering feature can be trivially computed as $LS/N$.
By the algebraic identity $\Var(X){=}E[X^2]{-}E[X]^2$, BIRCH computes the variance
of a clustering feature as $\Var(X){=}\frac1N SS{-}(\frac1N\sum_i LS_i)^2$.
We will discuss the numerical problems with this approach in \refsec{sec:birchnumeric}.

A new data sample $x$ can be easily integrated into the clustering feature
using $\CFBI {+} \vec{x} {=} (\vec{LS} {+} \vec{x}, SS {+} \sum_i x_i^2, N {+} 1)$.
Because all of these are sums, two clustering features can also easily be combined
(c.f., additivity theorem in~\cite{DBLP:conf/sigmod/ZhangRL96})
$\CFBI_A {+} \CFBI_B {=} (\vec{LS}_A {+} \vec{LS}_B, SS_A {+} SS_B, N_{\!A} {+} N_{\!B})$.
A single data point $\vec{x}$ can hence be interpreted as the clustering feature
containing $(\vec{x},\sum_i x_i^2, 1)$.

\subsection{Clustering Feature Tree (CF-Tree)}
\label{sec:CFTree}
The cluster features are organized in a depth-balanced tree called CF-Tree.
A leaf stores a set of clustering features (each representing one or many data points),
while the inner nodes store the aggregated clustering features
of each of its children.
The tree is built by sequential insertion of data points
(or, at a rebuild, the insertion of earlier clustering features).
The insertion leaf is found by choosing the ``nearest'' clustering feature at each level
(five different definitions of closeness will be discussed in \refsec{sec:measures}).
Within the leaf node, the data point is added to the best
clustering feature if it is within the merging ``threshold'',
otherwise a new clustering feature is added to the leaf.
Leaves that exceed a maximum capacity are split, which can propagate to higher levels of the
tree and cause the tree to grow when the root node overflows.
If the tree exceeds the memory limit, a new tree is built
with an increased merging threshold  by reinserting
the existing clustering features of the leaf level.
After modifying a node, the aggregated
clustering features along the path to the root are updated.

The discussion of BIRCH in textbooks ends with the CF-Tree,
although we do not yet have clusters.
This is because the outstanding idea of BIRCH is that of data aggregation into
clustering features, and we can run different clustering algorithms afterward.
The BIRCH authors mention hierarchical clustering,
$k$-means, and CLARANS \cite{DBLP:journals/tkde/NgH02}. %
For best results, we would want to use an algorithm
that not only uses the mean of the clustering feature, but that also uses the
weight and variance. The weight can be fairly easily used in many algorithms,
but the variance is less obvious to integrate.
In \refsec{sec:betulagmm} we will propose how to perform
Gaussian Mixture Modeling and use the variance information.

\subsection{BETULA Cluster Features}
\label{sec:betula-cf}

The way variance is computed from BIRCH cluster features 
using the popular equation $\Var(X){=}E[X^2]{-}E[X]^2$
is prone to the numerical problem known as ``catastrophic cancellation''.
This equation can return zero for non-constant data, and because of
rounding even negative values (and hence, undefined standard deviation).
In the context of BIRCH, we cannot resort to the numerically more reliable
textbook definition for variance, $\Var(X){:=}\frac1N\sum (x{-}\mu)^2$,
because this requires two passes over the data set (one to find $\mu$, then one for $\Var$).
But we also cannot just ignore the problem, because not all clustering features
will be close to $0$, where the numerical accuracy is not a problem.
Schubert and Gertz~\cite{DBLP:conf/ssdbm/SchubertG18}
discuss methods to compute variance and covariance for weighted data,
which forms the base for our approach.
For this, they collect three running statistics, very similar to
the three components of BIRCH clustering features:
(i) the sum of weights, (ii) the weighted mean (centroid vector),
and (iii) the weighted sum of squared deviations from the mean.
Clearly (i) corresponds to $N$ in the clustering feature,
(ii) is equivalent to $LS/N$, but (iii) is $\SSE{:=}\sum_x \weight_x \snorm{x{-}\mu}^2$
(where $\weight_x$ is the weight of the data point, often simply 1).
Hence, we propose the following replacement cluster feature for BETULA:
\begin{align}
\CFBE&:=(\weight, \mu, \SSE)
\end{align}
where $\weight$ is the aggregated weight of all data points
(BETULA also allows for weighted data samples),
$\mu$ denotes the current mean vector, and
$\SSE$ is the sum of squared deviations from the mean.
The last component can either be a scalar value as in BIRCH (the sum over all components)
or a vector of squared deviations. For our experiments, we chose the latter option;
a similar modification to BIRCH can be found in various publications
(e.g.,~\cite{DBLP:conf/vldb/AggarwalHWY03,DBLP:conf/vldb/AggarwalHWY04,DBLP:conf/sdm/CaoEQZ06,DBLP:journals/kais/KranenABS11,DBLP:books/mk/HanKP2011}).
A~single data point of weight $\weight_x$ is equivalent to a cluster feature
$\CFBE_x{=}(\weight_x,x,0)$ (because it has zero deviation from the mean).
Similar to the additivity theorem of BIRCH, we can efficiently combine two BETULA cluster features into one:
\begin{align}
\weight_{\!\AB} &= \weight_{\!A} + \weight_{\!B}
\label{eq:update-w}
\\
\mu_{\!\AB} &= \mu_{\!A} + \tfrac{\weight_{\!B}}{\weight_{\!\AB}} (\mu_{\!B}-\mu_{\!A})
\label{eq:update-mu}
\\
\SSE_{\!\AB} &= \SSE_A + \SSE_B + \weight_{\!B} (\mu_{\!A}-\mu_{\!B})(\mu_{\!\AB}-\mu_{\!B})
\label{eq:update-sse}
\end{align}
The derivation of these equations follows directly from the update equations for
the weighted (co-) variance of \cite{DBLP:conf/ssdbm/SchubertG18}.
Because their experiments indicate that using the sum of squared deviations,
$\SSE{:=}\sum_x n_x(x{-}\mu)^2$ has slight computational advantages, we follow suit.
We could also have stored $\Var{=}\SSE{/}\weight$ instead (we did not
measure a noticeable performance difference between these two options).

\subsection{Distance and Absorption Measures}
\label{sec:measures}
The BIRCH algorithm uses two different measures during tree construction.
The first is a distance between two clustering features, which is used to find the
closest leaf in the tree. The second is an absorption criterion, used together with a
threshold to decide when to add the new data to an existing or as a new clustering feature.
Both measures can be defined on the original data, but also in terms of the clustering
feature values to compute them efficiently.

\paragraph{Distance Measures:}
BIRCH proposes five different distance measures enumerated as D0 to D4.
The first two correspond simply to the Euclidean distance of the centers (D0)
and the Manhattan distance of the centers (D1).
The third, average inter-cluster distance (D2), is based on the
quadratic mean distance between points of different clusters,
while the average intra-cluster distance (D3) uses the quadratic mean
distance within the combined cluster.
Variance-increase distance (D4) is the variance of the resulting cluster
minus the variance of the separated clusters.
Similar ideas can be found in hierarchical clustering: centroid linkage (D0, D1),
average linkage (D2, D3), and Ward linkage (D4).
\begin{align}
\text{D0}(A,B) &= \snorm{\mu_{\!A}-\mu_{\!B}}_{\phantom{1}} = \sqrt{\textstyle\sum_i (\mu_{A,i}-\mu_{B,i})^2}
\label{eq:D0}
\\
\text{D1}(A,B) &= \snorm{\mu_{\!A}-\mu_{\!B}}_1 = \kern10pt \textstyle\sum_i |\mu_{A,i}-\mu_{B,i}|
\label{eq:D1}
\\
\text{D2}(A,B) &= \sqrt{\tfrac1{\weight_{\!A}\weight_{\!B}}\textstyle\sum_{x \in A} \sum_{y \in B}\snorm{x - y}^2}
\label{eq:D2}
\\
\text{D3}(A,B) &= \sqrt{\tfrac1{\weight_{\!\AB}(\weight_{\!\AB} - 1)}\textstyle\sum_{x, y \in \AB} \snorm{x - y}^2}
\label{eq:D3}
\\
\text{D4}(A,B) &= \sqrt{\textstyle
\sum_{x \in \AB}\snorm{x-\mu_{\!\AB}}^2
-\!\sum_{x \in A}\snorm{x-\mu_{A}}^2
-\!\sum_{x \in B}\snorm{x-\mu_{B}}^2}
\label{eq:D4}
\end{align}

\paragraph{Absorption Criteria:}
Absorption in BIRCH is based on a second criterion and a threshold.
Conceptually, the threshold can be seen as a maximum radius of a cluster feature;
if adding a point would increase the radius beyond the allowed maximum,
a new cluster feature is created instead of merging.
Intuitively, the radius should be defined as $\max_x \snorm{x{-}\mu}$;
but this value cannot be efficiently computed from the summary statistics.
Instead, the ``radius'' can be approximated using different criteria.
In BIRCH, these criteria were defined on a single clustering feature $AB$;
they are computed by virtually merging two clustering features and
evaluating the criteria on the result.
We can easily remove the distinction between distance and absorption criteria,
but one may nevertheless want to choose them differently
(e.g., choosing the nearest leaf
by Euclidean distance, but thresholding on minimum variance), as they serve a
different purpose.
The first criterion proposed in BIRCH is called ``radius'' R (Eq. \refeq{eq:R}),
the second is the ``diameter'' D (Eq. \refeq{eq:D}).
Both this ``radius'' and ``diameter'' are not maximum values, but averages:
the average distance to the center is (R), and the average distance of any two
points is (D), which happens to be the same as $D3(A,B)$.
Many implementation attempts (such as sklearn's) of BIRCH simply use the distance between the two
cluster centers instead (E) -- this cannot be defined in the original BIRCH
architecture but is easy to add.

\begin{align}
\text{R}(AB) &=\kern-7ex& \text{R}(A,B) &= \sqrt{\tfrac{1}{\weight_{\!\AB}}\textstyle\sum_{x \in \AB} \snorm{x - \mu_{\!\AB}}^2}
\label{eq:R}
\\
\text{D}(AB) &=\kern-7ex& \text{D3}(A,B) &= \sqrt{\tfrac{1}{\weight_{\!\AB}(\weight_{\!\AB} - 1)}\textstyle\sum_{x, y \in \AB} \snorm{x - y}^2}
\label{eq:D}
\\
\text{E}(A,B) &=\kern-7ex& \text{D0}(A,B) &= \snorm{\mu_{\!A} - \mu_{\!B}}
\label{eq:E}
\end{align}
The values of D and R are almost identical (use Eq.~\refeq{eq:verschiebung} below):
they differ only by a factor of $\frac{2n}{n{-}1}$; similar to the regular radius and diameter.
Because of the way they are used in BIRCH, we cannot expect them to perform very differently.
\subsection{Catastrophic Cancellation in BIRCH}
\label{sec:birchnumeric}
The numerical problem in BIRCH
arises from the ``textbook'' equation
for variance, $\Var(X){=}E[X^2]{-}E[X]^2$. This equation---while mathematically correct---is prone to
catastrophic cancellation when used with floating-point arithmetic,
unless $E[X]^2{\ll}E[X^2]$ holds \cite{DBLP:conf/ssdbm/SchubertG18}.
In clustering, we cannot assume that all clusters are close to the origin,
and the ideal leaves have a small variance and represent the data by their differences in the mean.
Because of this, it may not be sufficient to center the data globally.
Furthermore, we do not know the center beforehand, and in BIRCH we only want to do a single pass over the data for performance.

Unfortunately, both of the original absorption criteria R and D,
as well as distance measures D2--D4 are computed using
above ``textbook'' equality
\begin{align}
\Var(X) =&
\textstyle
\frac{1}{2n^2}
\tsum_{x,y \in X}\snorm{x-y}^2
= \tfrac{1}{n}
\textstyle
\tsum_{x\in X}\snorm{x-\mu_X}^2
\label{eq:verschiebung}
\intertext{which yields the following equalities for BIRCH (equivalent to $n{\cdot}\kern-2pt\Var(X){=}S$)}
\textstyle
S=&
\tsum_{x\in X}\snorm{x-\mu_X}^2
= 
\tfrac{1}{2n}
\textstyle
\tsum_{x\in X}\snorm{x-y}^2
=
SS - \tfrac{1}{n}\snorm{LS}^2.
\label{eq:sse}
\end{align}

The BIRCH authors hence proposed to compute these measures
(we omit D0, D1, and E as they do not involve squares)
based on clustering features as:
\begin{align}
\text{D2}(A,B) &= \sqrt{\tfrac{1}{N_{\!A} N_{\!B}}(N_{\!B} SS_A + N_{\!A} SS_B \badminus 2 LS_A^T LS_B)}
\label{eq:d2-birch}
\\
\text{D3}(A,B) &= \sqrt{\tfrac{2}{N_{\!A}+N_{\!B}-1}(SS_A+SS_B \badminus \tfrac{1}{N_{\!A}+N_{\!B}}\snorm{LS_A + LS_B}^2)}
\\
\text{D4}(A,B)
&=\sqrt{\tfrac1{N_{\!A}}\!\snorm{LS_A}^2 + \tfrac1{N_{\!B}}\!\snorm{LS_B}^2
\badminus\tfrac1{N_{\!A}+N_{\!B}}\!\snorm{LS_A + LS_B}^2}
\label{eq:d4-birch}
\\
\text{R}(AB) &= \sqrt{\tfrac{1}{N_{\!\AB}}(SS_{\!\AB}\badminus\tfrac{1}{N_{\!\AB}}\snorm{LS_{\!\AB}}^2)}
\\
\text{D}(AB) &= \sqrt{\tfrac{2}{N_{\!\AB}-1}(SS_{\!\AB}\badminus\tfrac{1}{N_{\!\AB}}\snorm{LS_{\!\AB}}^2)}
\end{align}
The subtractions flagged with a warning symbol \Warning{} can
suffer from catastrophic cancellation and hence numerical problems.
It may come unexpected that in the ``variance increase'' equation (D4)
all SS terms cancel out, and we only get the vector product of the linear sums,
but this is %
the Konig-Huygens theorem.

The effect of the catastrophic cancellation usually leads to an underestimation
of the actual variance, and hence of the distances. Because of this, data points
may be assigned to the wrong branch or node. While the result will not be
completely off, it is easy to avoid these problems in the first place.
More severe problems arise when using the resulting variance in the subsequent
steps, such as in clustering. Because most implementations of BIRCH
only use the centers of the leaf entries for clustering (e.g., sklearn
does not even use the weight, and only supports Euclidean distance D0),
this has not been observed frequently.

Much of the later work based on BIRCH is prone to the same problem
in one way or another. In CF-kernel density estimation \cite{DBLP:conf/kdd/ZhangRL99},
the variance is bounded to guarantee approximation quality --
underestimating the variance invalidates this guarantee.
The (diagonal) Mahalanobis distance used in \cite{DBLP:conf/kdd/BradleyFR98}
divides by the standard deviation, which can become 0 due to instabilities;
the division tends to amplify the errors.
CluStream~\cite{DBLP:conf/vldb/AggarwalHWY03} uses the standard deviation
of the arrival times, estimated with the unstable equation.
HPStream~\cite{DBLP:conf/vldb/AggarwalHWY04} relies on
per attribute standard deviations for subspace clustering.
DenStream \cite{DBLP:conf/sdm/CaoEQZ06} uses the radius $R$
to estimate density, while data bubbles \cite{DBLP:conf/pkdd/BreunigKS00}
rely on the standard deviation to estimate the extent.
ClusTree \cite{DBLP:journals/kais/KranenABS11} estimates the
variance in this unstable way.
All of these methods can easily be modified to use BETULA cluster features.

Using the improved BETULA cluster features introduced in \refsec{sec:betula-cf},
which we will simply denote by $\CF$, 
we can easily avoid these numerical problems, because these features directly
aggregate the squared errors instead of the sum of squares,
as previously used for online estimation of variance \cite{DBLP:conf/ssdbm/SchubertG18}.

\subsection{Improved Distance Computations}

In BETULA cluster features,
we use the mean $\mu$ instead of the linear sum because this makes the subsequent
operations more efficient (and elegant). The update equations
for merging \CF{}s also involve the mean (c.f.~Eq.~\refeq{eq:update-sse}),
and we can now compute the BIRCH distances in a more numerically stable way.
Using BETULA cluster features $\CF{=}(\weight, \mu, \SSE)$,
and \refeqn{eq:sse}, we can compute the
distances and absorption criteria now as follows (the derivation is
included in the arXiv draft):
\begin{align}
\text{D0}(A,B) &= \snorm{\mu_{\!A}-\mu_{\!B}}
\\
\text{D1}(A,B) &= \snorm{\mu_{\!A}-\mu_{\!B}}_1
\\
\text{D2}(A,B) &= \sqrt{\tfrac{1}{\weight_{\!A}}\SSE_A + \tfrac{1}{\weight_{\!B}}\SSE_B +\snorm{\mu_{\!A}-\mu_{\!B}}^2}
\\
\text{D3}(A,B) &= \sqrt{\tfrac{2}{\weight_{\!\AB}(\weight_{\!\AB}-1)}(\weight_{\!\AB} (\SSE_A+\SSE_B)+\weight_{\!A}\weight_{\!B} \snorm{\mu_{\!A}-\mu_{\!B}}^2)}
\\
\text{D4}(A,B) &= \sqrt{\tfrac{\weight_{\!A}\weight_{\!B}}{\weight_{\!\AB}}\snorm{\mu_{\!A}-\mu_{\!B}}^2}
\\
\text{R}_{\!\AB} &= \sqrt{\tfrac{1}{\weight_{\!\AB}}\SSE_{\!\AB}}
= \sqrt{\tfrac{1}{\weight_{\!\AB}}(\SSE_A + \SSE_B + \tfrac{\weight_{\!A} \cdot \weight_{\!B}}{\weight_{\!\AB}} \snorm{\mu_{\!A} -\mu_{\!B}}^2)}
\\
\text{D}_{\!\AB} &= \sqrt{\tfrac{2}{(\weight_{\!\AB} - 1)}\SSE_{\!\AB}}
= \text{D3}(A,B)
\end{align}
With these numerically more stable equations, we can build a CF-Tree using BETULA cluster features
instead of the original BIRCH clustering features.

\subsection{Gaussian Mixture Modeling with BETULA Cluster Features}
\label{sec:betulagmm}
Gaussian Mixture Modeling (GMM) with the EM algorithm \cite{journals/jroyastatsocise2/DempsterLR77} is
a popular, but fairly expensive clustering algorithm. Every iteration, the probability density functions (pdfs)
of each Gaussian are evaluated at every data point, then the distribution parameters are updated based
on all points weighted by their probabilities. Because this is a soft clustering, a
tolerance threshold or an iteration limit are used for convergence. Formally, the method is linear in the number
of data points, but in practice, it is fairly expensive because of the many
pdfs to compute and the number of iterations. To scale this algorithm to large data sets (large~$n$)
as well as many clusters~$k$, it is beneficial to use a data summarization technique
such as BIRCH or BETULA.
Several variations of GMM exist:
we can restrict cluster shapes and we can have independent or shared model parameters.
MAP estimation can be employed to improve the robustness~\cite{DBLP:journals/classification/FraleyR07},
because there are other numerical pitfalls that can lead
to degenerate clusters.
We only consider some of the more popular variants in this work:
the spherical model with varying weight and identical volume in each dimension (IGMM),
the diagonal model with varying weight and different volume in each dimension (DGMM),
and the fully variable model that models covariance (CGMM). 
If we only have a scalar for $SS$, then this is well-suited for the simplest
model: A spherical model, in which the direction of variance does not matter.
When using a vector for $\SSE$, we can incorporate this per-axis information into
the cluster models. For the arbitrarily oriented model, we would need to use a
covariance in each cluster feature. This is possible using the
corresponding equations for the covariance of \cite{DBLP:conf/ssdbm/SchubertG18},
but the memory requirement increases to $1{+}d{+}\binom{d}{2}=1{+}\frac{d(d{+}1)}{2}$
values per cluster feature. Because of this, we do not include this in the experiments.

For clustering, the main tree structure is usually discarded, and only the
cluster features within the leaf nodes are kept. For the initialization of the
algorithm, we apply the kmeans++~\cite{DBLP:conf/soda/ArthurV07} initialization
on the leaf entries.
Afterward, the Gaussian Mixture Modeling algorithm is executed.

In classic GMM, we usually process a single data sample
at a time. When processing cluster features, these represent multiple objects.
To improve the quality of the clustering, rather than just using the cluster mean to represent a Cluster Feature,
we use the Gaussian distribution of the data in the \CF{}, which we assume is better
(at least for GMM).
To estimate the responsibilities of each cluster for each clustering we then use $\int_x \mathcal{N}(x|\mu_1,\sigma^2_1)\mathcal{N}(x|\mu_2,\sigma^2_2)dx = \mathcal{N}(\mu_1|\mu_2,\sigma^2_1{+}\sigma^2_2)$.
Using the law of total probability, these values are normalized to sum to 1,
exactly as in the usual EM procedure. When updating the cluster models,
the weight of the cluster features is trivially usable as additional weight,
and we can update the model variance using \refeqn{eq:update-sse}. 

By utilizing BETULA cluster features and EM-GMM it is possible to cluster big
data sets with limited memory and high numerical stability as shown
in \refsec{sec:eval}. It is also possible to distribute this
procedure into a cluster by partitioning the data and aggregating the models
of all nodes (c.f.~\cite{DBLP:conf/ssdbm/SchubertG18}).

\section{Evaluation}
\label{sec:eval}

We compare the following alternative implementations of GMM:
\begin{texttabbing}
\hspace{\parindent}\=\hspace{1.6cm}\=\kill
\> Textbook \> Standard EM~\cite{journals/jroyastatsocise2/DempsterLR77} using the equation $E[X^2]{-}E[X]^2$
\\
\> Stable \> Numerically stable EM implementation (from ELKI \cite{DBLP:journals/corr/abs-1902-03616,DBLP:conf/ssdbm/SchubertG18})
\\
\> BIRCH \> EM-style using the original BIRCH clustering features
\\
\> BETULA \> EM-style using our new BETULA cluster features
\end{texttabbing}

\noindent
The evaluation of clustering algorithms is inherently difficult
because they are used in an unsupervised context, where no labeled
data is available. Real data is usually dirty and contains
undesirable artifacts (such as anomalies, duplicate values, and
discretization effects) that can cause problems for
methods that assume continuous data. GMM is
no exception: e.g., constant attributes will break
many implementations.
In these experiments, we do not aim at showing the superiority of
Gaussian Mixture Modeling over other approaches. The limitations
of it are well understood, in particular when data has non-convex
clusters.

Instead, we focus on the following research questions:
\begin{texttabbing}
\hspace{1cm} \= \kill
RQ1 \> How numerically (un-)stable is BIRCH, does BETULA help?
\\
RQ2 \> Is the quality of BETULA comparable with BIRCH and regular GMM?
\\
RQ3 \> How does BETULA scale with data set size (and compare to BIRCH)?
\\
RQ4 \> Are the results applicable to real data?
\end{texttabbing}

\subsection{Experimental Setup}
We modify the existing implementations of BIRCH and GMM clustering of
ELKI 0.7.5 \cite{DBLP:journals/corr/abs-1902-03616}.
By keeping most of the code shared, we try to
minimize the effects caused by implementation differences,
as recommended for comparing algorithms
\cite{DBLP:journals/kais/KriegelSZ17}.
All computations are executed on a small cluster with Intel E5-2697v2 CPUs,
we do not use multithreading, and we repeated each experiment 10 times with varying random seeds and data input order,
and give the average results.
All the CF-Trees are built using the variance-increase distance (D4, Eq.~\ref{eq:D4})
in combination with the radius absorption criterion (R, Eq.~\ref{eq:R}).
This combination yields subclusters with low variance as input for the GMM clustering.
We do not present results with other distances and absorption criteria here because of redundancy;
they were similar.
The size of CF-Trees is by default limited to 5000 leaf entries unless specified differently;
when this number is exceeded the tree is rebuilt with a bigger threshold as in BIRCH.
For the GMM clustering step, all algorithms are initialized by kmeans++~\cite{DBLP:conf/soda/ArthurV07}.
After 100 iterations or when no further improvement can be made the optimization is stopped.

\subsection{Numerical Stability}

\begin{figure}[b!]
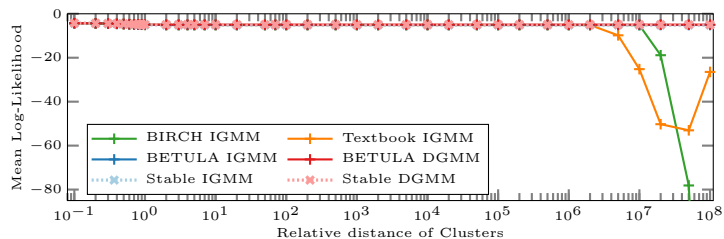
\centering
\plotShiftLike{\linewidth/1.2}{\linewidth/3}
\caption{The log-likelihood goodness of fit of the model with increasing distance between
the clusters demonstrates the numerical instability of some algorithms.}
\label{fig:shift_like}
\end{figure}

First, we demonstrate the numerical instability using synthetic data with two Gaussian clusters
in $\mathbb{R}^3$ of 150000 points.
Both clusters have standard deviations $[\tfrac{4}{3}, 1, \tfrac{3}{4}]$,
and the only variable in the test is how far the clusters are shifted away from the mean.
For small separation, both clusters overlap but with increasing distance, the clustering gets trivial
until numerical stability comes into play.

The impact of the increasing distance between the clusters can be seen in \reffig{fig:shift_like}
where all algorithms provide good results until first the Textbook IGMM implementation
at $5{\cdot}10^6$ and then BIRCH IGMM at $2{\cdot}10^7$ begin to deteriorate.
The degeneration of BIRCH IGMM begins a bit later than Textbook IGMM
because of the aggregation in the CF-Tree helping a bit, but it then fails even worse.
A deterioration at $10^7$ is to be expected from double-precision because of the squared values;
with single-precision floating-point, it is to be expected to occur at a separation of $10^3$.
Both the ``Stable'' regular GMM and BETULA are not affected and solve this idealized toy
problem without difficulties (RQ1).

\subsection{Quality Comparison on Synthetic Data}

\begin{figure}[tb]
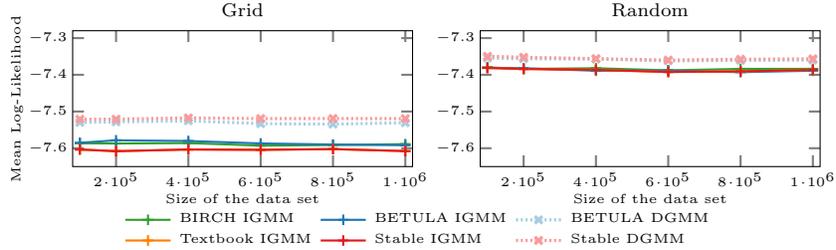
\centering
\plotCombLike{\linewidth}{\linewidth/2.4}
\caption{Log-Likelihood goodness of fit of the model on both synthetic data sets.}
\label{fig:combined_like}
\end{figure}

We now address the question of result quality (RQ2) in a scenario where all algorithms are stable.
For the evaluation, two synthetic data sets are used,
which are similar to data used for the evaluation of the original
BIRCH algorithm \cite{DBLP:conf/sigmod/ZhangRL96} but larger and with increased variability.
We use the data generator of ELKI~\cite{DBLP:journals/corr/abs-1902-03616},
which has a convenient size multiplier parameter for this experiment.

The first data set is called \enquote{Grid} and consists of a 10 by 10 grid of clusters
with a distance of 5 between the means of the clusters on each axis.
Each cluster consists of 10000 points with a variance per attribute randomly drawn from $\mathcal{N}(1,0.25)$.
The second data set, \enquote{Random}, consists of 100 clusters in a 50~by~50 area with the
cluster means distributed by Halton sampling, which produces a pseudo-random uniform distribution.
The variance of each cluster is again specified by a normal distribution $\mathcal{N}(1,0.15)$.
This time the size of each cluster varies and is randomly drawn from between 5000 and 15000 points.

\reffig{fig:combined_like} shows the log-likelihood of the models on these data sets.
For both, it can be seen that the data set size has next to no influence on the quality of the fit.
The models with diagonal variance (Stable DGMM and \mbox{BETULA} DGMM) produce a better fit than the models that are
restricted to using the same variance in each attribute.
On the \enquote{Random} data set,
all IGMM approaches perform similar (as expected).
On the \enquote{Grid} data set, both BETULA IGMM and BIRCH IGMM unexpectedly
achieve a higher likelihood than the standard IGMM algorithms.
This difference can be explained by the fact that the implementations using cluster features
converge faster (because there are fewer objects) than the approaches that use the raw data;
the latter have not yet converged within the maximum number of iterations.
However, this experiment is designed to test if BETULA performs similar to BIRCH on
the same test data that the BIRCH publications used, and to detect programming errors.

\subsection{Runtime Evaluation on Synthetic Data}
\begin{figure}[tb]
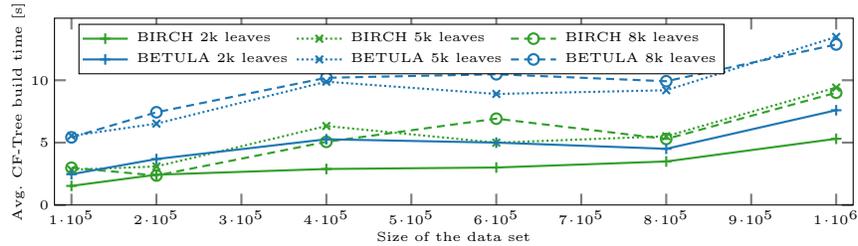
\centering
\plotRandomBuild{\linewidth}{\linewidth/3}%
\caption{Build time with a varying number of leaf entries on the random data set.}%
\label{fig:random_build}
\end{figure}

When evaluating the runtime of BETULA with GMM clustering two measurements are of interest: 
The time to build the CF-Tree only, and the time for the entire clustering procedure.
\reffig{fig:random_build} shows the time BETULA and BIRCH need to build the CF-Tree for various tree sizes.
It can be seen that the time for building the tree increases with the size of the data set
and also with the size of the tree due to an increasing number of distance calculations for the insertion of new points.
The construction time for BETULA is larger than for BIRCH because this
implementation uses a vector for storing the variances, while the BIRCH implementation
uses only a scalar;
but the tree construction is only a small part of the total time as we will see next.

\begin{figure}[tb]
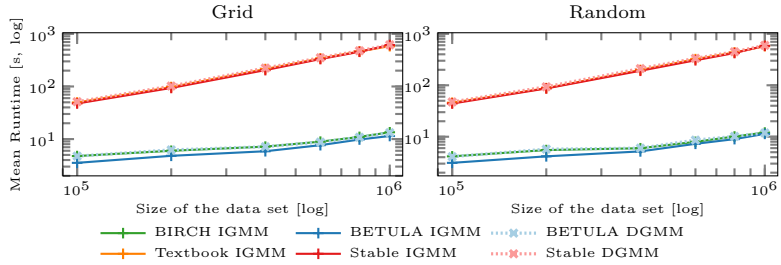
\centering
\plotCombTime{\linewidth}{\linewidth/3.5}%
\caption{Runtime of the clustering on both synthetic data sets.}%
\label{fig:combined_time}
\end{figure}

When looking at the complete runtime of BIRCH (respectively BETULA) including GMM clustering,
shown in \reffig{fig:combined_time}, we can see that the standard GMM algorithms have
a much higher runtime by a factor of 18 to 52 on this data set, due to the compression
achieved by the CF-Tree (which improves with data set size).
We use a log-log plot to see the differences between BIRCH and BETULA,
which perform very similar (RQ3).
BETULA is up to 5\% faster than BIRCH---despite using a vector to store variances---%
because the BETULA cluster features can be used directly
for clustering, while more additional computations are necessary with
BIRCH clustering features to obtain mean and variance on the fly.

\subsection{Clustering Real Data}
To test the algorithm on real data,
we use the location information of the UK ``Road Safety Data'' from 1979 to 2004 from
data.gov.uk.\footnote{https://data.gov.uk/dataset/cb7ae6f0-4be6-4935-9277-47e5ce24a11f}
This data set has about 6.2~million entries and contains data on road accidents from Great Britain.
The location information in this data set is given in the OSGR grid reference system which is only used in Great Britain;
which we convert to the appropriate UTM coordinate system.
For this experiment, we reduced the cluster feature precision from double precision to single precision
in both BIRCH and BETULA to demonstrate the numerical instabilities on real data.
The regular GMM clustering is performed with double precision to get a more precise reference value.

\begin{figure}[tb]
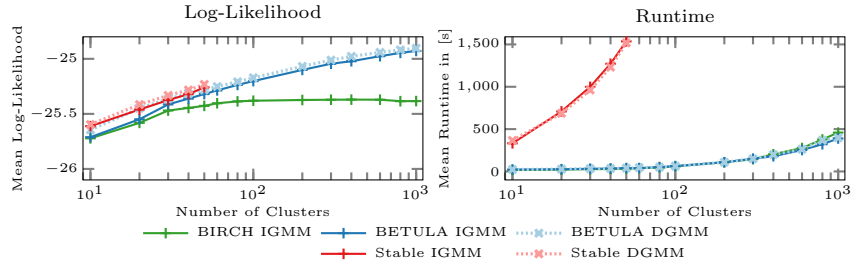
\centering
\plotTrafficCombined{\linewidth}{\linewidth/3.5}
\caption{Log-Likelihood goodness of fit of the models and runtime on the traffic accident data with 15000 leaf entries
(Stable GMM only up to 50 clusters because of runtime).}
\label{fig:traffic_combined}
\end{figure}

\reffig{fig:traffic_combined} shows that Stable DGMM and Stable IGMM achieve a better fit
to the data than the CF-Tree based approximations (which is to be expected, given that
they use the individual points and double precision).
However, the runtime of this method is much higher, and hence it is only computed up
to $k{=}50$ clusters.
BETULA with DGMM and IGMM clustering obtain only slightly worse results,
showing that the BETULA cluster features provide a reasonably close approximation of the data.
BIRCH IGMM on the other hand shows its numerical instabilities and with an increasing number
of clusters, the quality deteriorates compared to BETULA.
For numerous clusters (and a large value makes sense on this data set),
BETULA with DGMM delivers the best results at an acceptable run time:
As seen in \reffig{fig:traffic_combined}, all GMM with Stable, BIRCH, and BETULA
scale approximately linear in the number of clusters $k$;
but because the CF-Trees reduce the data set from 6.2 million to at most
15000 cluster features (a factor of over 400), we obtain good results
at a much smaller run time than with regular Stable DGMM or IGMM (RQ4).

\begin{figure}[tb!]\centering
\includegraphics[height=.45\linewidth]{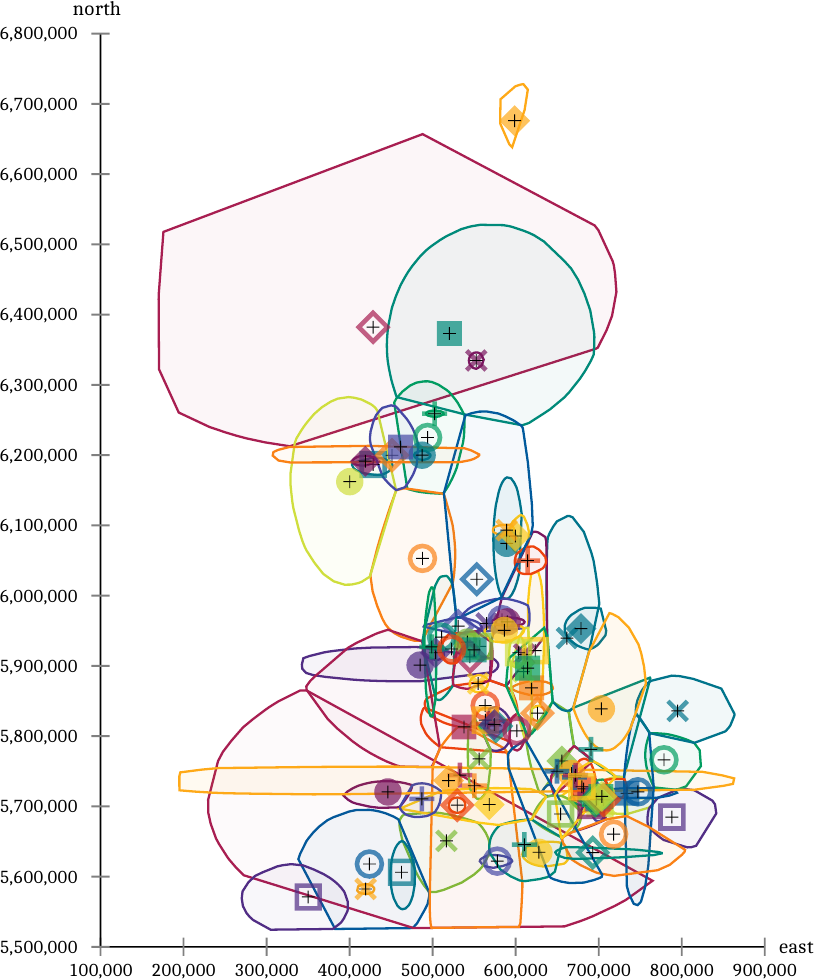}
\caption{Convex hulls of clusters with BETULA DGMM on the traffic accident data with 100~clusters
(three clusters omitted for a cleaner visualization).}
\label{fig:traffic_shape}
\end{figure}

\reffig{fig:traffic_shape} shows the convex hulls and cluster centroids of
an exemplary clustering of the traffic data set with $k{=}100$ clusters,
using BETULA DGMM and visualized with ELKI.
We removed three clusters containing only input data errors to improve readability.
The shape of Great Britain can be recognized; small and dense clusters are found around the larger
British cities, especially London.
Larger clusters with lower density on the other hand cover rural areas with fewer accidents
(it is typical behavior of GMM to nest dense clusters with low variance inside ``background'' clusters
with high variance and fewer data points).

\section{Conclusion}

Big data analysis and data stream clustering are hot topics in today's research.
The CF-Tree of BIRCH is a popular technique for
this that inspired many subsequent works.
Recently, the reliability of machine learning is receiving increased attention;
unfortunately, we found that ``catastrophic cancellation'' is a major
problem when calculating variances in BIRCH and derived methods,
which can cause the results to deteriorate.

In this article, we proposed BETULA cluster features, that can serve
as a drop-in replacement for BIRCH. These do no longer exhibit this
problem as they avoid using the unstable equation, at a negligible
performance difference. We also show how to use BETULA to accelerate
Gaussian Mixture Modeling, while using the variance information from
the cluster features for improved quality, compared to the standard
approach of only using the centroids of each leaf entry.

\bibliographystyle{splncs04}
\bibliography{Chapter/VarianceClustering}

\appendix
\section{Derivation of the BETULA Distance Criteria}
The derivations rely on the following two equivalences:
\begin{align*}
\tsum_{i=1}^{n}\tsum_{j=1}^{n}\snorm{x_i{-}x_j}^2 = 2n\tsum_{i=1}^{n}\snorm{x_i{-}\mu}^2
\\
\SSE_{AB} = \SSE_A + \SSE_B + \tfrac{\weight_A\weight_B}{\weight_A+\weight_B}\snorm{\mu_A{-}\mu_B}^2
\end{align*}

We can now derive the following equations to compute the distances using BETULA cluster features.
First, average inter-cluster distance:
\begin{align*}
\text{D2}(A,B)^2 &= \tfrac1{\weight_{\!A}\weight_{\!B}}\tsum_{x \in A} \tsum_{y \in B}\snorm{x{-}y}^2
\\
&= \tfrac1{2\weight_{\!A}\weight_{\!B}} \Big(\tsum_{x,y \in AB}\snorm{x{-}y}^2 - \tsum_{x,y \in A}\snorm{x{-}y}^2 - \tsum_{x,y \in B}\snorm{x{-}y}^2 \Big)
\\
&= \tfrac1{\weight_{\!A}\weight_{\!B}}\Big(\weight_{\!AB}\tsum_{x\in AB} \snorm{x{-}\mu}^2 - \weight_{\!A}\tsum_{x\in A} \snorm{x{-}\mu}^2 - \weight_{\!B}\tsum_{x\in B} \snorm{x{-}\mu}^2\Big)
\\
&= \tfrac1{\weight_{\!A}\weight_{\!B}}\left(\weight_{\!AB}\SSE_{AB} - \weight_{\!A}\SSE_A - \SSE_B\right)
\\
&= \tfrac1{\weight_{\!A}\weight_{\!B}}\left(\weight_{\!AB}\left(\SSE_A + \SSE_B + \tfrac{\weight_A\weight_B}{\weight_A+\weight_B}\snorm{\mu_A{-}\mu_B}^2\right) - \weight_{\!A}\SSE_A - \SSE_B\right)
\\
&= \tfrac{\weight_{B}}{\weight_{\!A}\weight_{\!B}} \SSE_A + \tfrac{\weight_{A}}{\weight_{\!A}\weight_{\!B}} \SSE_B + \tfrac{\weight_A\weight_B}{\weight_A\weight_B}\snorm{\mu_A{-}\mu_B}^2
\\
&= \tfrac{1}{\weight_{\!A}}\SSE_A + \tfrac{1}{\weight_{\!B}}\SSE_B +\snorm{\mu_{\!A}{-}\mu_{\!B}}^2
\shortintertext{Second, average intra-cluster distance:}
\text{D3}(A,B)^2 &= \tfrac1{\weight_{\!\AB}(\weight_{\!\AB} - 1)}\tsum_{x, y \in \AB} \snorm{x{-}y}^2
= \tfrac2{\weight_{\!\AB}(\weight_{\!\AB} - 1)} \weight_{AB} \tsum_{x \in AB}\snorm{x{-}\mu}^2
\\
&= \tfrac2{\weight_{\!\AB}(\weight_{\!\AB} - 1)} \weight_{AB} \SSE_{AB}
\\
&= \tfrac2{\weight_{\!\AB}(\weight_{\!\AB} - 1)} \weight_{AB} \left( \SSE_A + \SSE_B + \tfrac{\weight_A\weight_B}{\weight_{AB}}\snorm{\mu_A{-}\mu_B}^2 \right)
\\
&= \tfrac{2}{\weight_{\!\AB}(\weight_{\!\AB}-1)}(\weight_{\!\AB} (\SSE_A+\SSE_B)+\weight_{\!A}\weight_{\!B} \snorm{\mu_{\!A}{-}\mu_{\!B}}^2)
\shortintertext{Third, variance-increase distance:}
\text{D4}(A,B)^2 &= \tsum_{x \in \AB}\snorm{x{-}\mu_{\!\AB}}^2
-\!\tsum_{x \in A}\snorm{x{-}\mu_{A}}^2
-\!\tsum_{x \in B}\snorm{x{-}\mu_{B}}^2
\\
&= \SSE_{AB}-\SSE_A-\SSE_B
= \SSE_A + \SSE_B + \tfrac{\weight_A\weight_B}{\weight_{AB}} \snorm{\mu_A{-}\mu_B}^2-\SSE_A-\SSE_B
\\
&= \tfrac{\weight_{\!A}\weight_{\!B}}{\weight_{\!\AB}}\snorm{\mu_{\!A}{-}\mu_{\!B}}^2
\shortintertext{Last, radius (average distance to the center):}
\text{R}(A,B)^2 &=  \tfrac{1}{\weight_{\!\AB}}\tsum_{x \in \AB} \snorm{x{-}\mu_{\!\AB}}^2
= \tfrac{1}{\weight_{\!\AB}}\SSE_{\!\AB}
\\
&= \tfrac{1}{\weight_{\!\AB}}(\SSE_A + \SSE_B + \tfrac{\weight_{\!A} \cdot \weight_{\!B}}{\weight_{\!\AB}} \snorm{\mu_{\!A}{-}\mu_{\!B}}^2)
\end{align*}

We do not provide derivations for Euclidean distance $D0$, Manhattan distance $D1$, as these
can trivially be computed from the cluster means $\mu$ using the standard definitions of these distances.
The diameter $D$ is equivalent to $D3$.

\section{Derivation of the BIRCH Distance Criteria}
The derivations rely on the following equivalence:
\begin{align*}
\tfrac{1}{n} \tsum_{i=1}^{n}\snorm{x{-}\mu}^2 &= \Var(X^2) = E(X^2)-E(X)^2
= \tfrac{1}{n}SS - \snorm{\tfrac{LS}{n}}^2
\end{align*}

Using this---numerically problematic---equivalence we can reproduce the equations
used in the original BIRCH algorithm \cite{tr/wisc/Zhang97}:
\begin{align*}
\text{D2}(A,B)^2 &= \tfrac1{\weight_{\!A}\weight_{\!B}}\tsum_{x \in A} \tsum_{y \in B}\snorm{x{-}y}^2
\\
&= \tfrac1{2\weight_{\!A}\weight_{\!B}} \Big(\tsum_{x,y \in AB}\snorm{x{-}y}^2 - \tsum_{x,y \in A}\snorm{x{-}y}^2 - \tsum_{x,y \in B}\snorm{x{-}y}^2 \Big)
\\
&= \tfrac1{\weight_{\!A}\weight_{\!B}}\Big(\weight_{\!AB}\tsum_{x\in AB} \snorm{x{-}\mu}^2 - \weight_{\!A}\tsum_{x\in A} \snorm{x{-}\mu}^2 - \weight_{\!B}\tsum_{x\in B} \snorm{x{-}\mu}^2\Big)
\\
&= \tfrac1{\weight_{\!A}\weight_{\!B}}\Big( (\weight_{AB}SS_{AB} - \snorm{LS_{AB}}^2)-(\weight_{A}SS_{A} - \snorm{LS_{A}}^2)-(\weight_{B}SS_{B} - \snorm{LS_{B}}^2)\Big)
\\
&= \tfrac{1}{n_{\!A} n_{\!B}}\left(n_{\!B} SS_A + n_{\!A} SS_B + (\snorm{LS_{A}}^2+ \snorm{LS_{B}}^2 - \snorm{LS_{A} +LS_{B}}^2)\right)
\\
&= \tfrac{1}{n_{\!A} n_{\!B}}(n_{\!B} SS_A + n_{\!A} SS_B - 2 LS_A^T LS_B)
\\
\text{D3}(A,B)^2 &= \tfrac1{\weight_{\!\AB}(\weight_{\!\AB} - 1)}\tsum_{x, y \in \AB} \snorm{x{-}y}^2
= \tfrac{2\weight_{AB}}{\weight_{\!\AB}(\weight_{\!\AB} - 1)}\tsum_{x \in AB}\snorm{x{-}\mu}^2
\\
&= \tfrac{2}{\weight_{\!\AB} - 1} \left(SS_{AB}- \weight_{AB}\snorm{\tfrac{LS_{AB}}{\weight_{AB}}}^2\right)
\\
&= \tfrac{2}{n_{\!A}+n_{\!B}-1}(SS_A+SS_B - \tfrac{1}{n_{\!A}+n_{\!B}}\snorm{LS_A{+}LS_B}^2)
\\
\text{D4}(A,B)^2 &= \tsum_{x \in \AB}\snorm{x{-}\mu_{\!\AB}}^2
-\!\tsum_{x \in A}\snorm{x{-}\mu_{A}}^2
-\!\tsum_{x \in B}\snorm{x{-}\mu_{B}}^2
\\
&= SS_{AB}-\tfrac{1}{\weight_{AB}}\snorm{LS_{AB}}^2 -\left( SS_{A}-\tfrac{1}{\weight_{A}}\snorm{LS_{A}}^2 \right) -\left( SS_{B}-\tfrac{1}{\weight_{B}}\snorm{LS_{B}}^2 \right)
\\
&= SS_{AB} - SS_{A} - SS_{B} + \tfrac{1}{\weight_{A}}\snorm{LS_{A}}^2 + \tfrac{1}{\weight_{B}}\snorm{LS_{B}}^2 - \tfrac{1}{\weight_{AB}}\snorm{LS_{AB}}^2
\\
&=\tfrac1{n_{\!A}}\!\snorm{LS_A}^2 + \tfrac1{n_{\!B}}\!\snorm{LS_B}^2 -\tfrac1{n_{\!A}+n_{\!B}}\!\snorm{LS_A{+}LS_B}^2
\\
\text{R}(A,B)^2 &= \tfrac{1}{\weight_{\!\AB}}\tsum_{x \in \AB} \snorm{x{-}\mu_{\!\AB}}^2
=\tfrac{1}{\weight_{\!\AB}}SS_{AB}-\snorm{\tfrac{LS_{AB}}{n_{AB}}}^2
\\
&=\tfrac{1}{n_{\!\AB}}(SS_{\!\AB}-\tfrac{1}{n_{\!\AB}}\snorm{LS_{\!\AB}}^2)
\end{align*}

\end{document}